\documentclass[letterpaper, 10 pt, conference]{llncs}

\usepackage[breaklinks,hidelinks,bookmarksnumbered,pdfencoding=auto,psdextra,bookmarksdepth=1]{hyperref} 
\usepackage{graphics} 
\usepackage{times} 
\usepackage{amsmath} 
\usepackage{amssymb} 
\usepackage{graphicx}
\usepackage[binary-units]{siunitx} 
\usepackage{wrapfig}
\usepackage{tabularx}
\usepackage{booktabs} 
\usepackage{makecell}
\usepackage{multirow} 
\usepackage{hhline}
\usepackage{enumitem}

\usepackage[ruled, linesnumbered, vlined, noend]{algorithm2e}

\SetCommentSty{mycommfont}
\usepackage[noend]{algpseudocode}

\usepackage{tikz}
\usetikzlibrary{backgrounds, fit, shapes.geometric, positioning}
\usepackage{url}
\usepackage{color,soul}
\usepackage{rotating}

\usepackage[font=footnotesize]{caption} 
\usepackage[font=footnotesize, labelfont=footnotesize, textfont=footnotesize, labelformat=brace]{subcaption} 

\makeatletter 
\renewcommand\p@subfigure{\thefigure\,}
\makeatother 

\def\secref#1{Sec.~\ref{#1}}
\def\figref#1{Fig.~\ref{#1}}
\def\subfigref#1{\subref{#1})}

\def\tabref#1{Tab.~\ref{#1}}

\def\eqref#1{Eq.~(\ref{#1})}
\def\algref#1{Alg.~\ref{#1}}
\def\alglineref#1{line~\ref{#1}}
\algnewcommand{\algorithmicgoto}{\textbf{go to}}
\algnewcommand{\Goto}[1]{\algorithmicgoto~line~\ref{#1}}

\newcommand\etal{\emph{et al.}}
\newcommand{\comment}[1]{}

\newcolumntype{Y}{>{\centering\arraybackslash}X}
\begin{document}
\mainmatter 
\title{Reactive Correction of Object Placement Errors\\for Robotic Arrangement Tasks} 
\titlerunning{Reactive Correction of Object Placement Errors}
\toctitle{Reactive Correction of Object Placement Errors\\for Robotic Arrangement Tasks}

\author{Benedikt Kreis \and Rohit Menon \and Bharath Kumar Adinarayan \and\\ Jorge de Heuvel \and Maren Bennewitz
}%
\institute{Humanoid Robots Lab, University of Bonn, 53115 Bonn, Germany\\ \email{\{kreis,menon,adinarayan,deheuvel,maren\}@cs.uni-bonn.de} } 

\authorrunning{B. Kreis \etal} 

\maketitle
\thispagestyle{empty} 
\pagestyle{empty}

\begin{abstract}
When arranging objects with robotic arms, the quality of the end result strongly depends on the achievable placement accuracy.
However, even the most advanced robotic systems are prone to positioning errors that can occur at different steps of the manipulation process.
Ignoring such errors can lead to the partial or complete failure of the arrangement.
In this paper, we present a novel approach to autonomously detect and correct misplaced objects by pushing them with a robotic arm.
We thoroughly tested our approach both in simulation and on real hardware using a Robotiq two-finger gripper mounted on a UR5 robotic arm.
In our evaluation, we demonstrate the successful compensation for different errors injected during the manipulation of regular shaped objects.
Consequently, we achieve a highly reliable object placement accuracy in the millimeter range.

\end{abstract} 

\section{Introduction}
\label{sec:intro}

Object assembly is one of the most complex manipulation skills for both humans and machines. Assembly in the industrial workplace has slowly progressed from using machines with fixtures and jigs to using robots with vision and other sensors.  
However, while industrial assembly is a challenging task, it consists of parts that were meant to be assembled according to a plan and it takes place in a structured environment. 
Reconstruction of historical relics using robots, on the other hand is a much harder challenge as it consists of the assembly of fragile yet priceless frescoes, pottery, and statues that are discovered as fragments at archaeological sites.
Currently, most of the research regarding reconstruction of ancient relics focuses on virtual reconstruction~\cite{pintusSurveyGeometricAnalysis2016}.
Physical reconstruction in the real world is even more challenging as precise placement planning is needed and only finger grippers can be used to avoid touching the painted fragment surfaces.

Research on placement planning has received less attention in contrast to grasp planning till now, as grasping is an essential skill required to complete many real-world tasks and it has to be performed before placing an object.
The most common use case of grasping is bin picking where individual objects are grasped and then dropped or tossed into boxes according to the assigned object class without regards to the final placement pose~\cite{kleebergerSurveyLearningBasedRobotic2020}.
However, if there are manipulation constraints with respect to other objects as in an assembly, the accurate final placement pose is crucial, e.g., when building towers~\cite{furrerAutonomousRoboticStone2017} or aligning two objects to fit a screw~\cite{vondrigalskiUncertaintyAwareManipulationPlanning2022}.

Independent of the manipulated object, any robotic system is prone to errors.
There are internal and external factors that influence the manipulation.
Internal factors include the manufacturing precision of mechanical parts, sensor/actuator noise, the used control strategy, and the selected motion planning algorithms.
External factors include humans working in the vicinity of the robot and environmental conditions.

\begin{figure}[t]
	\centering
	\includegraphics[width=\textwidth]{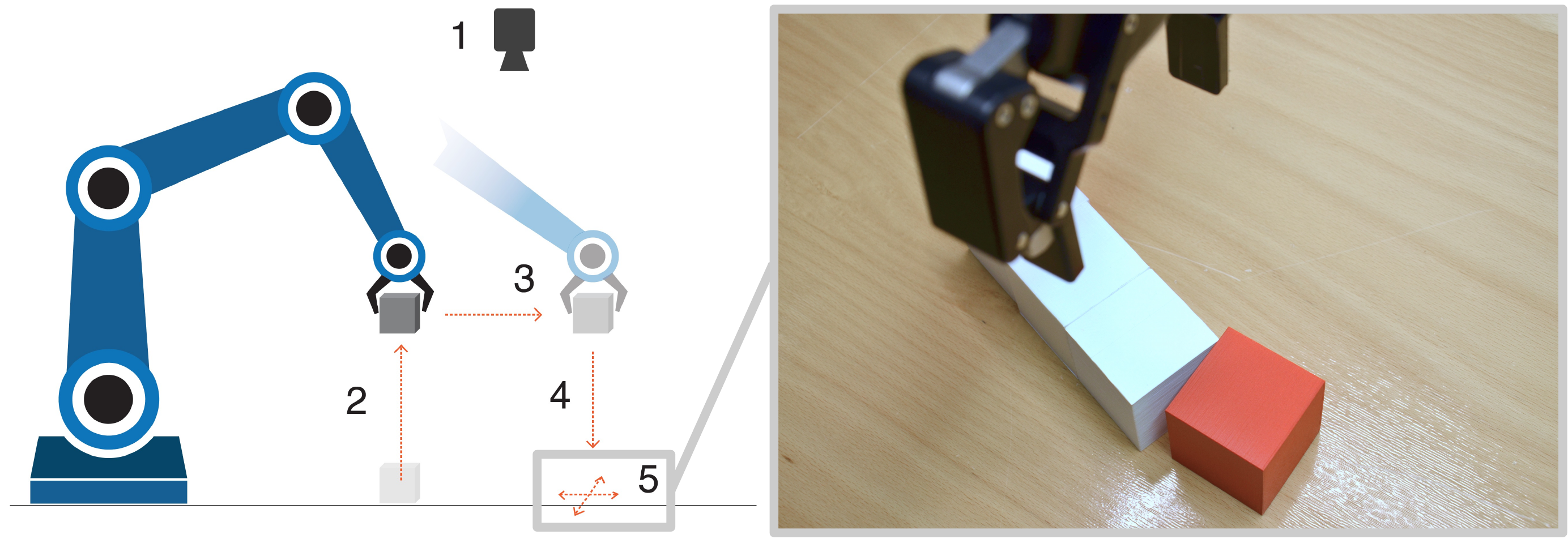}
	
	\vskip 0.5\baselineskip
	
	\begin{minipage}[c]{.49\textwidth}
		\centering
		\subcaption{Manipulation process}
		\label{fig:process}
	\end{minipage}
	\begin{minipage}[c]{.49\textwidth}
		\centering
		\subcaption{Arrangement example}
		\label{fig:arrangement}
	\end{minipage}
	\vskip -\baselineskip
	\caption{\subfigref{fig:process} The object
	manipulation process using a robotic arm can be divided into five steps:
	1)~Perception, 2)~Picking, 3)~Transport, 4)~Placement,
	5)~Correction. \subfigref{fig:arrangement} The goal of the last step is to correct
	misplaced objects such as the red cube.}
	\label{fig:motivation}
\end{figure}

In this paper, we focus on the task of placing objects according to a structured arrangement plan.
In this plan, each object has an assigned placement pose, but it is not intended to physically connect the objects with each other.
The plan is executed by picking and placing the objects as shown in steps 1 to 4 in \figref{fig:process}.
First, the object is detected and its pose estimated.
Then, a grasp pose is generated.
In the third step, the object is transported to the desired location and placed in the fourth step.
We assume that at this point placement errors exist due to aforementioned factors, which cannot be completely eliminated.
Therefore, we propose to add a fifth step to the standard pick and place pipeline in which the object's pose is verified and reactively corrected if needed.

We implemented our approach using a robotic arm with a two-finger gripper that performs picking, placing, and pushing of objects.
In order to focus on the core problem of misplaced objects and not on complex object recognition, we use cubes similar to the regular shaped objects of Fu \etal~\cite{fu6DRoboticAssembly2022}, Nägele \etal~\cite{nageleLegoBotAutomatedPlanning2020}, and Wei \etal~\cite{weiVisionguidedFineoperationRobot2021}.
We tested our approach in simulation and then successfully transferred it to a real robotic platform.
In both  scenarios, we show that manipulation errors can be minimized by iteratively correcting misplaced objects.
This is especially important in the real world, where these errors can have a great impact on the manipulation process.
The main contributions of this paper are: 

\begin{itemize}[label=\textbullet] 
	\item The systematic categorization of errors that influence the placement of objects.
	\item A novel approach to reactively correct objects given their estimated misplacement.
	\item The experimental evaluation of our approach including the sim-to-real transfer.
\end{itemize} 

\section{Related Work}
\label{sec:related}

As one of the most complex problems in robotics, assembly planning presents challenges in task planning, sensing, perception, manipulation, motion planning, and actuation. State-of-the-art approaches tend to focus on a single aspect of assembly. However, with current trends towards personalization and customization of products, researchers are increasingly focusing on the planning and execution of complete assembly plans based on product specifications while taking robot capabilities into account \cite{rodriguezIterativelyRefinedFeasibility2019}. 

Perception is a crucial component of assembly planning for estimating object poses. While earlier computer vision approaches used to focus on template matching and CAD-model-based pose estimation, recent approaches have taken advantage of deep learning networks to estimate poses with six degrees of freedom (6-DoF) for known objects from RGB images alone, or combined with depth images \cite{berscheidRobotLearningDoF2021a},~\cite{labbeCosyPoseConsistentMultiview2020b},~\cite{wangGDRNetGeometryGuidedDirect2021a},~\cite{xiangPoseCNNConvolutionalNeural2018}.

However, training deep learning models for detecting different objects in the real world is laborious and computationally expensive. Moreover, quick inference requires large compute resources without any guarantee of the final pose estimation accuracy under different environmental conditions.
Hence, model-free approaches that combine perception and manipulation to estimate grasp poses have also been investigated~\cite{kleebergerSurveyLearningBasedRobotic2020}.
Berscheid \etal~\cite{berscheidSelfSupervisedLearningPrecise2020} used self-supervised learning to enable a robot to learn the complete pick and place task by showing a single RGB-D image of the desired goal state.  

In the field of construction robotics, Furrer \etal~\cite{furrerAutonomousRoboticStone2017} showed that it is possible to autonomously stack two to four irregularly shaped lime stones on top of each other using a robotic arm with a 3-finger gripper.
The authors presented a full pipeline of stone detection, feasible pose searching, and object manipulation.
By using a physics engine and a simulation they are able to find suitable stacking poses which the real robot uses to put the stones on top of each other.
Nevertheless, placement errors and the overall tower stability are not considered which can result in the collapse of the tower.

Approaches that try to reduce the placement pose error actively are still rare.
Fu \etal~\cite{fu6DRoboticAssembly2022} tried to cope with placement errors by performing a so-called 3-axis calibration to center the object below the gripper before performing the final assembly step.
But this last step consists of another pick and place operation which again poses the risk of introducing errors. 
Zhao \etal~\cite{zhaoPreciseRoboticGrasping2021} proposed to predict the post displacement accuracy in order to select an adequate grasp which is expected to result in the minimal error.
However, during the execution of the pick and place unpredictable errors can occur so that they cannot guarantee a certain placement accuracy.
Von Drigalski \etal~\cite{vondrigalskiUncertaintyAwareManipulationPlanning2022} propose the use of so-called "static" (touch, look) and “extrinsic” actions (grasp, place, push) that are used to reduce the object's pose uncertainty by exploiting gravity, the environment, and the gripper geometry.
But the authors limit their planner to generating action sequences of a maximum length of three actions while we constrain our correction actions to the desired placement accuracy using a threshold which can be set depending on the application.
Furthermore, our approach minimizes the placement error in an iterative refinement step in contrast to Zhao \etal~\cite{zhaoPreciseRoboticGrasping2021}, Berscheid \etal~\cite{berscheidSelfSupervisedLearningPrecise2020}, Fu \etal~\cite{fu6DRoboticAssembly2022}, and Furrer \etal~\cite{furrerAutonomousRoboticStone2017},
who perform the placement action in an open loop manner.
This allows us to eliminate errors which can occur during the previous manipulation steps.

It is common to use abstracted models to study assembly problems such as colored building blocks (Fu \etal~\cite{fu6DRoboticAssembly2022}), LEGO structures (Nägele \etal~\cite{nageleLegoBotAutomatedPlanning2020}), or puzzles (Wei \etal~\cite{weiVisionguidedFineoperationRobot2021},~\cite{weiVisionGuidedHandEye2021}).
This way, methods and algorithms can be developed and evaluated focusing on core problems, rather than on complex object detection and pose estimation.
That is why we use color-coded cubes with identical dimensions in this work.

\section{Occurrence of Manipulation Errors}
\label{sec:errors}
\begin{figure}[t]
	\centering
	\includegraphics[width=\textwidth]{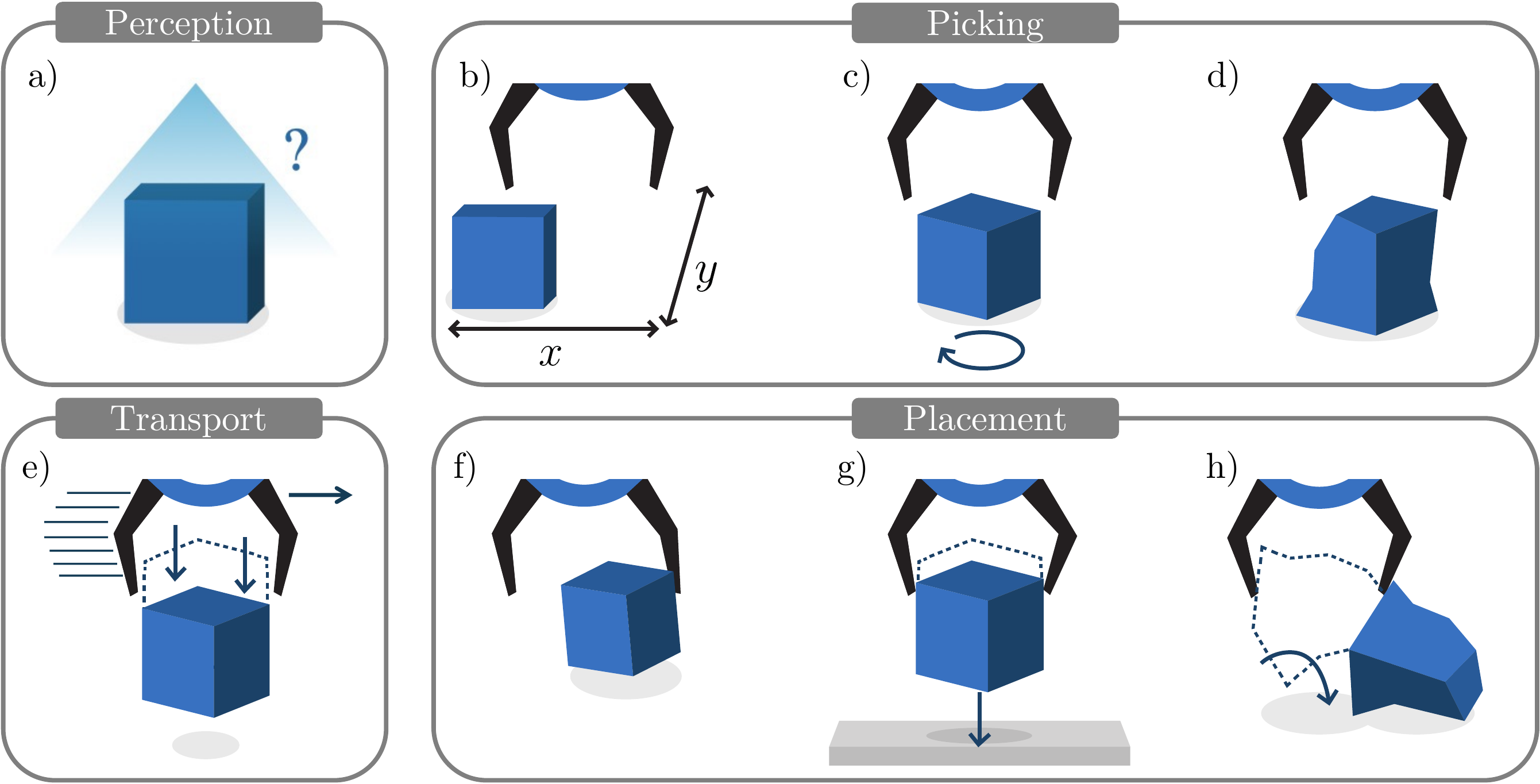}
	\caption{Placement errors can occur during the entire manipulation process: a) The object recognition or pose estimation are incorrect. b) The gripper is not centered above the object. c) The gripper fingers are not aligned to the object surface. d) The object's surface is uneven. e) The object slips during transport. f) The object sticks to one finger while opening the gripper. g) The object is released above the table to avoid collision. h) The object moves after opening the gripper.}
	\label{fig:manipulationErrors}
\end{figure}

It is important to distinguish between different manipulation errors because their influence on the object's placement pose varies.
We focus on errors that occur during the pick and place process and that significantly influence the accuracy of the placement pose.
The identified errors illustrated in \figref{fig:manipulationErrors} are divided into four categories according to their order of occurrence during the pick and place process.
Note that errors may be propagated through different process steps.
Hence, the categories do not reflect the error source but the point when they affect the manipulation.

\begin{enumerate}[label=\bfseries \arabic*)] 
  \item \textbf{Perception}
\begin{itemize}[label=\textbullet] 
    \item \textbf{Object detection and pose estimation error:}
It is a challenging task to detect objects due to the amount of possible dimensions, structures, and colors.
Also light conditions and occlusions can play a vital role.
Failing to overcome these challenges can result in the incorrect segmentation of the image and as a consequence also in the faulty estimation of the object's pose (\figref{fig:manipulationErrors}\,a).
\end{itemize} 
  \item \textbf{Picking}
\begin{itemize}[label=\textbullet] 
    \item \textbf{Translation error:}
When the gripper is not fully centered above the object, the fingertips drag it over the table until they are closed, which leads to an offset between the planned and actual object pose after the grasp (\figref{fig:manipulationErrors}\,b).
This may be due to pose estimation errors or insufficient robot accuracy.
    \item \textbf{Orientation error:}
If the gripper and object surface are not perfectly aligned, a parallel grip leads to orientation errors while grasping it (\figref{fig:manipulationErrors}\,c).
    \item \textbf{Surface error:}
Objects can have an irregular surface so that the contact area between the gripper and the object differs depending on the grasp pose.
Therefore, when grasping the object it may change its pose (\figref{fig:manipulationErrors}\,d).
\end{itemize} 
  \item \textbf{Transport}
\begin{itemize}[label=\textbullet] 
    \item \textbf{Grip error:}
Objects may slip in the gripper during transport from the pick to the place location (\figref{fig:manipulationErrors}\,e).
This can be due to jerk or acceleration forces from the executed robot movement.
Also insufficient contact and unsuitable material pairings of gripper and object can result in an unintended drop.
\end{itemize} 
  \item \textbf{Placement}
\begin{itemize}[label=\textbullet] 
    \item \textbf{Friction error:}
Depending on the combination of gripper and object materials, the object may stick to one gripper finger during the placement (\figref{fig:manipulationErrors}\,f).
    \item \textbf{Release error:}
In order to avoid a crash between the table and a manipulated object, it is a common practice at production lines to drop the object a few millimeters above the desired placement position which potentially leads to pose errors (\figref{fig:manipulationErrors}\,g).
    \item \textbf{Tilting error:}
Objects may have an irregular bottom which means they can change their pose when opening the gripper.
In the worst case, they can tip over if their center of mass is unfavorably distributed (\figref{fig:manipulationErrors}\,h).
\end{itemize} \end{enumerate}

Some of the errors can be reduced by taking mechanical measures such as adapting the material of the gripper fingers to the manipulated objects.
However, there may not be a suitable solution for all manipulated objects, so that misplacements still occur.

\section{Reactive Correction of Placement Errors}
\label{sec:approach}

The goal of our work is to find and evaluate an effective correction procedure for objects that were misplaced due to the errors identified in \secref{sec:errors}.
We propose to add a fifth step to the standard pick and place pipeline in which the object's pose is verified and reactively corrected if needed (see \figref{fig:process}). We assume that the perception is accurate enough to allow for an effective correction. The required perception accuracy depends on the application.  

\subsection{Overview}
\label{subsec:process}

\algref{algo:flow} shows an overview of our approach for \(n\) objects.
First the object is detected in the scene and its pose is estimated which is used to pick the object.
Then it is transported to and placed at the desired location.
Subsequently, the pose estimation is performed again to calculate the offset.
This serves as an input for the correction step which is repeated to iteratively refine the placement pose until a threshold is reached.
This threshold is defined by the accuracy of all the components involved in the manipulation process and it is quantified in the experimental evaluation (\secref{sec:evaluation}).

\begin{algorithm}[t]
	\SetAlgoLined
	\DontPrintSemicolon
	\SetNoFillComment
	
	\SetKwInOut{KwInput}{Input}
	\SetKwInOut{KwOutput}{Output}	
	\KwInput{Arrays of object identifiers \(\mathit{objectIDs}\) and placement poses \(\mathit{desiredPoses}\).}
	\KwOutput{Array of offsets \(\mathit{offsets}\) with remaining misplacements.}
	{	\(\mathit{n} \gets 0\)\;
		\While{\( \mathit{n} < \mathit{size}(\mathit{objectIDs}) \)}
		{	\(\mathit{success} \gets \mathit{False}\)\;
			\(\mathit{objectID} \gets \mathit{objectIDs}[\mathit{n}]\)\;
			\(\mathit{desiredPose} \gets \mathit{desiredPoses}[\mathit{n}]\)\;
			\(\mathit{objectPose} \gets \mathit{getPose}(\mathit{objectID})\)\label{algo:flow:berscheid} \tcp*[l]{Step 1 in \figref{fig:process}}
			\(\mathit{pickObject}(\mathit{objectPose})\)\label{algo:flow:errorinjection} \tcp*[l]{Step 2 in \figref{fig:process}}
			\(\mathit{placeObject}(\mathit{desiredPose})\) \tcp*[l]{Steps 3 \& 4 in \figref{fig:process}}
			\(\mathit{objectPose} \gets \mathit{getPose}(\mathit{objectID})\)\;
			\(\mathit{offset} \gets \mathit{calculateOffset}(\mathit{desiredPose},\ \mathit{objectPose})\)\;
			\While{\(\mathit{offset} >= \mathit{threshold}\)\label{algo:flow:threshold}}
			{
				\(\mathit{pushObject}(\mathit{desiredPose},\ \mathit{objectPose})\)\label{algo:flow:correction} \tcp*[l]{Step 5 in \figref{fig:process}}
				\(\mathit{objectPose} \gets \mathit{getPose}(\mathit{objectID})\)\;
				\(\mathit{offset} \gets \mathit{calculateOffset}(\mathit{desiredPose},\ \mathit{objectPose})\)\;
			}
			\(\mathit{n} \gets \mathit{n}+1 \)\label{algo:flow:marker}\;
			\(\mathit{offsets}[\mathit{n}] \gets \mathit{offset}\)\;
		}
	}
	\caption{Manipulation with reactive correction for \(n\) objects}
	\label{algo:flow}
\end{algorithm}

\subsection{Iterative Correction}
\label{subsec:strategies}

During grasping the object gets automatically aligned to the gripper's fingertips so that the desired orientation for the grasp pose is achieved.
This happens due to the parallel closing mechanism of the gripper and its flat fingertips.
As a result, the orientation accuracy after placing the object is mainly dependent on the correct gripper orientation during grasping.
Since cubes perfectly align with the fingertips, and we initially orient the gripper according to the desired grasp pose, the orientation error after grasping is minimal.
Hence, we mainly focus on the translation error during the correction step.

We considered two correction strategies: Regrasping and pushing.
Regrasping means a repetition of manipulation steps 1 to 4 (see \figref{fig:process}), which poses the risk that the final object pose is affected by the same errors as in the first loop.
Pushing, on the other hand, is a planar, non-prehensile operation with less degrees of freedom than regrasping.
This reduces the risk of potential errors, e.g.,
the object cannot slip out of the gripper fingers and it cannot stick to the gripper fingers when they are opened.
At the same time, it is also faster and more accurate than picking and placing \cite{masonMechanicsPlanningManipulator1986}, thus making it a more suitable strategy to correct placement errors, especially for small correction distances.

The correction step is divided into inspection and pushing.
The goal of the inspection is to check whether a correction is required or not.
This is determined by the offset, which is the difference between the desired placement pose and the current object pose.
Using RGB images allows to calculate the offset with millimeter accuracy in \(x\) and \(y\) with respect to the desired placement coordinates.
If the inspection necessitates a correction, the object is iteratively pushed by the respective offset distance while alternating the direction in \(x\) and \(y\) between each push.
So we laterally push the object while the gripper opening is set to the width of the object minus the thickness of the fingers.
This way, we can effectively correct translation and small orientation displacements.

\section{Experimental Evaluation}
\label{sec:evaluation}

We implemented the pick, place, and push pipeline on a robotic platform to demonstrate the ability of our system to effectively reduce placement errors by reactively correcting object poses.
For this purpose, we use cubes with a uniformly colored top surface and an edge length of \SI{5}{cm}.
The robotic arm grasps the objects in the center and places them according to an arrangement plan on the table.
As an error metric for the placement accuracy of the cubes, we use the 2D~Euclidean distance \(\Delta d_{xy}\), which we measure with the top camera.
As in the work of Berscheid \etal~\cite{berscheidSelfSupervisedLearningPrecise2020}, the camera is calibrated and hence we do not require an external ground truth measurement system.

Since the occurrence and magnitude of manipulation errors as described in \secref{sec:errors} vary in a real arrangement of objects, we manually injected errors, which can occur, e.g., by using an erroneous grasp pose estimator.
Out of the presented errors, we focus on the translation error (\figref{fig:manipulationErrors}\,b) and the orientation error (\figref{fig:manipulationErrors}\,c), because they often appear during the crucial picking step.
However, our approach works for all errors elaborated in \secref{sec:errors}, assuming a well-calibrated hardware setup.
In simulation and in the real world, the errors are injected before grasping a cube by randomly moving it in \(x\), \(y\), and around \(z\).
This way, each cube was rotated up to \SI{\pm 40}{\degree} and shifted up to \SI{\pm 2.5}{cm}, which is half of its size.
These values correspond to the maximum planar displacement that still allows for a successful grasp of the cubes.
We verify that these errors actually appear in a real scenario by implementing an approach that produces such errors.
For this purpose, the object pose estimation step in \alglineref{algo:flow:berscheid} of \algref{algo:flow} is replaced with the machine-learning based 6-DoF grasp pose estimation from Berscheid \etal~\cite{berscheidRobotLearningDoF2021a}.
All experiments were performed in simulation as well as on the real robotic platform.
More details about our approach and experimental results are shown in the supplementary material\footnote{Videos:
\href{https://youtu.be/yt6Ct6JeoBs}{youtu.be/yt6Ct6JeoBs} (experiments), \href{https://doi.org/10.5281/zenodo.7925474}{doi.org/10.5281/zenodo.7925474} (presentation)
}.

\subsection{Experimental Setup}
\label{subsec:setup}

\begin{figure}[!ht]
	\centering
	\includegraphics[width=0.7\textwidth]{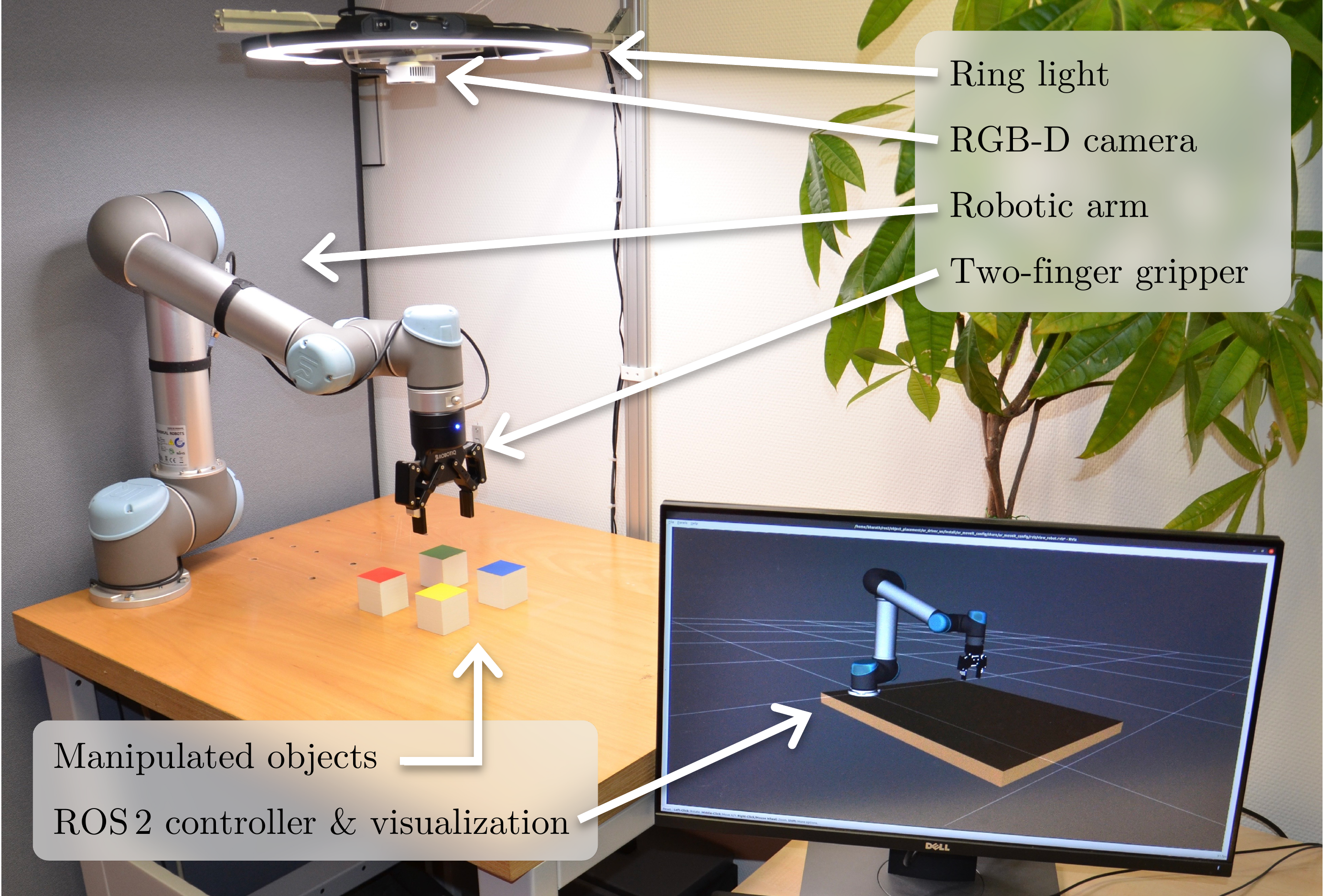}
	\caption{Our setup consists of several hardware components. We use a two-finger gripper mounted on a robotic arm. Both are controlled by a desktop computer. The cubes on the table are detected by an external RGB-D top camera. A ring light around the camera ensures adequate illumination.}
	\label{fig:hardwareSetup}
\end{figure}

We use the 6-DoF industrial robotic arm UR5 from Universal Robots mounted on top of a table to test our approach.
Attached to the tool center point is the two-finger gripper 2F-85 from Robotiq.
The Intel Realsense L515 RGB-D serves as a top camera (facing downwards).
It was fastened to a pole \SI{70}{cm} above the table.
A ring light installed around the camera ensures adequate illumination.
Furthermore, the setup includes a desktop computer equipped with an i9-11900KF CPU, 64\,GB RAM, and a RTX 3080Ti GPU with 12\,GB graphics RAM, which run the robot controller and visualization.
All components are shown in \figref{fig:hardwareSetup}.

We implemented our approach using the second version of the Robot Operating System (ROS\,2) as our robotic middleware \cite{macenskiRobotOperatingSystem2022}, and performed the simulation experiments using the physics simulator Gazebo Fortress \cite{koenigDesignUseParadigms2004a}.
Furthermore, we make use of MoveIt\,2~\cite{colemanReducingBarrierEntry2014} to plan and execute collision-free motions.

\subsection{Object Detection and Pose Estimation}
\label{subsec:detection}

We apply color thresholding on the RGB images from the top camera in the HSV color space to create a segmentation mask per cube.
A kernel of size 7\,\(\times\)\,7 applied to the mask corrects the noise using a closing operation (erosion followed by dilation) and an opening operation (dilation followed by erosion).
On the resulting mask we apply the algorithm of Suzuki and Abe~\cite{suzukiTopologicalStructuralAnalysis1985}, followed by the algorithm of Douglas and Peucker~\cite{douglasAlgorithmsReductionNumber1973} to estimate the cube's contour.
From the approximated contour we determine the four corners of all cube surfaces in the image plane and store them in a sorted list.

In order to calculate the 3D pose of the object with respect to the camera coordinate frame, we formulate a Perspective-n-Point problem using the previously computed cube corners, the known object dimensions, and the intrinsic camera parameters.
The solution is computed with a direct linear transformation, which is refined by a Levenberg-Marquardt optimization~\cite{moreLevenbergMarquardtAlgorithmImplementation1978}.
Using the extrinsic camera parameters, the algorithm transforms the object pose in camera coordinates C to world coordinates W:

\begin{equation}
\label{eq:transformation}
{}_\mathrm{W}^\mathrm{O}T =
{}_\mathrm{W}^\mathrm{C}T \cdot
{}_\mathrm{C}^\mathrm{O}T
\end{equation}

For the evaluation, we set the accuracy threshold of the final placement pose defined in \alglineref{algo:flow:threshold} of \algref{algo:flow} to \SI{1}{mm}.
Preliminary experiments showed that a lower threshold results in greater displacements than no correction at all.
This is due to system noise as well as the inertia of the object during the push and the friction with the table.

\subsection{Quantitative Evaluation of the Error Correction}
\label{subsec:correction}
\begin{figure}[t!]
	\newcommand{\picturewidth}{\textwidth}
	\newcommand{\cellwidth}{\textwidth}
	\begin{minipage}[c]{\textwidth}
		\centering
		\captionof{table}{Experimental results showing mean and standard deviation of the offset distances \(\Delta d_{xy}\)}
		\label{tab:results}
		\begin{tabularx}{\textwidth}{l*{4}{Y}}
			\toprule[\lightrulewidth]
			\multicolumn{1}{c}{\multirow{2}{*}{[mm]}}& \multicolumn{2}{c}{\textbf{a) Simulation}} & \multicolumn{2}{c}{\textbf{b) Real}} \\
			& After placing & After pushing & After placing & After pushing \\
			\midrule[\lightrulewidth]
			1) System accuracy & \(0.47 \pm 0.29\)  & \(0.40 \pm 0.25\) & \(3.72 \pm 2.30\) & \(0.72 \pm 0.34\) \\
			2) Translation error & \( 8.54 \pm 5.70\) & \(0.45 \pm 0.41\) & \(14.30 \pm 7.95\) & \(1.08 \pm 0.56\) \\
			3) Orientation error & \(3.03 \pm 4.70\) & \(0.43 \pm 0.25\) & \(4.18 \pm 3.74\) & \(0.80 \pm 0.39\) \\
			4) Berscheid \etal~\cite{berscheidRobotLearningDoF2021a} & \(5.18 \pm 3.83\) & \(0.41 \pm 0.25\) & \(14.18 \pm 10.72\) & \(1.07 \pm 0.50\) \\
			\bottomrule[\lightrulewidth]
		\end{tabularx}
	\end{minipage}
	\vskip 3.0\baselineskip
	\begin{minipage}[c]{\cellwidth}
		\centering
		\includegraphics[width=\picturewidth]{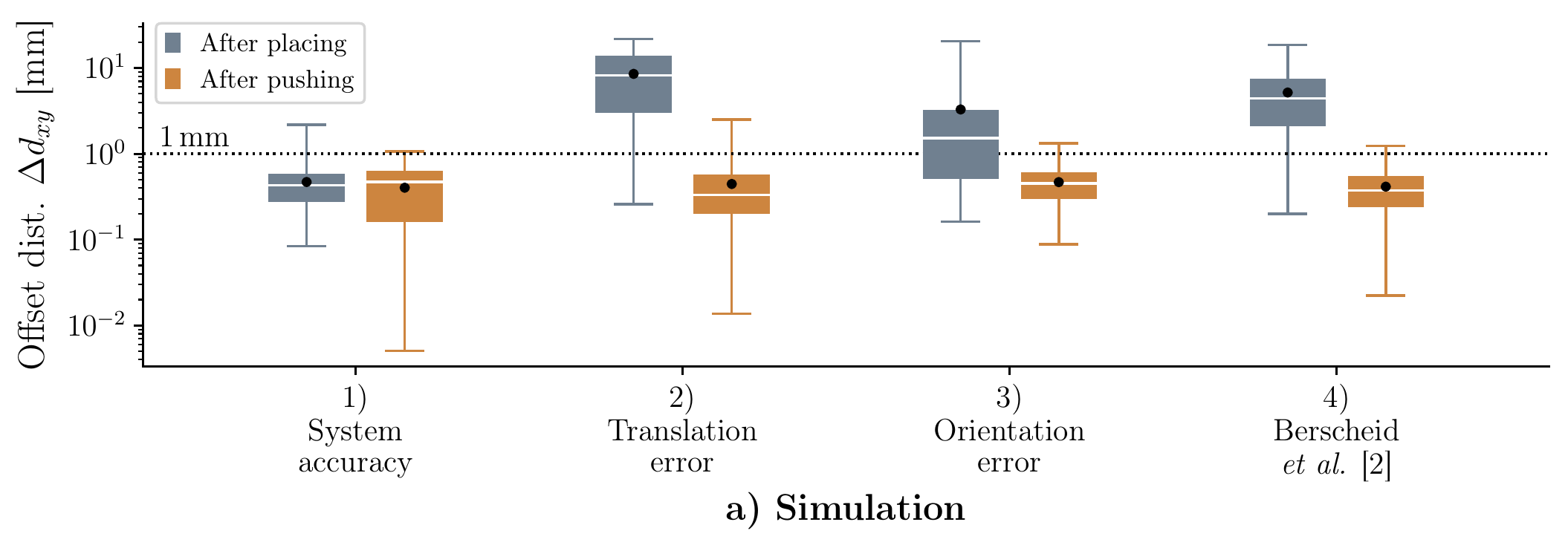}
	\end{minipage}
	\vskip 0.5\baselineskip
	\begin{minipage}[c]{\cellwidth}
		\centering
		\includegraphics[width=\picturewidth]{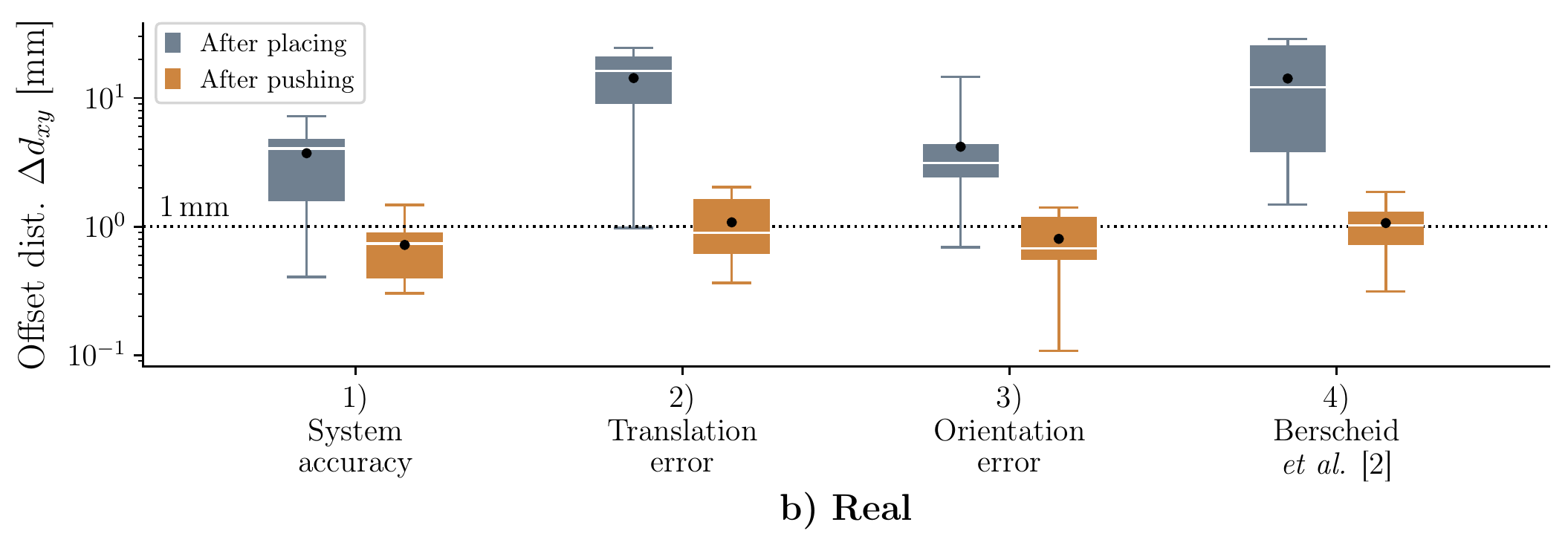}
	\end{minipage}	
	\caption{In the evaluation, we measured the offset distance \(\Delta d_{xy}\) (logarithmic vertical axis) between the desired and actual placement pose after placing and after pushing the object. The results are plotted as box plots where the white line represents the median and the black dot the mean. Four experiments were performed with 100 trials in simulation (a) and 10 trials in real world (b): 1) We determined the nominal system accuracy without injected error. 2) and 3) depict the accuracy with an injected translation and orientation error. 4) The approach of Berscheid \etal~\cite{berscheidRobotLearningDoF2021a} was used to estimate the grasp pose. As can be seen, the pushing effectively reduces the misplacement of objects in all experiments.}
	\label{fig:errorplots}
\end{figure}

The next set of experiments demonstrate that the reactive correction of misplaced objects lead to significantly more accurate placement poses.
In total, four sets of quantitative experiments were performed whose results are shown in \figref{fig:errorplots} and \tabref{tab:results}.
In all of them \algref{algo:flow} was used to execute pick, place, and push actions.
All experiments were executed 100 times in simulation (\figref{fig:errorplots}\,a) and 10 times with the real robot (\figref{fig:errorplots}\,b).
In the following paragraphs the term "experiment" refers to one row in \tabref{tab:results}.
Note that the effect of manipulation errors is greater in the real world than in simulation.
All measured mean offset distances \(\Delta d_{xy}\) after placing the object in the real world (\figref{fig:errorplots}\,b) are higher than in simulation (\figref{fig:errorplots}\,a).

\subsubsection{Nominal System Error:} 
\label{subsubsec:system}

In the first experiment (\figref{fig:errorplots}.1 and \tabref{tab:results}.1), the manipulation process was executed to evaluate the nominal system accuracy.
Using the estimated object pose, the desired grasp pose was selected such that the gripper fingertips are centered around the object edges.
\tabref{tab:results}.1 shows that there is an average error of \SI{0.47}{mm} in simulation and \SI{3.72}{mm} in the real world, before the correction. The error in the real-world experiments is almost a magnitude of order higher, thus revealing the negative impact of system noise, which is not simulated.
While pushing does not reduce the low error in simulation further, it leads to a significant improvement in final pose placement accuracy in the real world, with the error reducing by \(\sim\)\,\SI{80}{\percent} on an average to \SI{0.72}{mm}. 
At the same time, it demonstrates the need for reactive correction strategies to ensure a precise placement of objects.
Using an accuracy threshold of \SI{1}{mm}, this was achieved by just two corrections which demonstrates the efficiency of the pushing.

\subsubsection{Translation and Orientation Error:} 
\label{subsubsec:translationorientation}

In the second (\figref{fig:errorplots}.2 and \tabref{tab:results}.2) and third (\figref{fig:errorplots}.3 and \tabref{tab:results}.3) experiment, the translation error and the orientation error, respectively, were artificially injected before picking the object (before \alglineref{algo:flow:errorinjection} in \algref{algo:flow}), in order to simulate picking errors.
When comparing the data points after placing and after pushing the object in both experiments, a significant shift of the object's end position distribution towards the target location (horizontal axis) is visible.
This shows the increase of the placement accuracy after the correction step.
In simulation (\figref{fig:errorplots}\,a), this effect is even more visible than in the real world (\figref{fig:errorplots}\,b).
However, the most notable error reduction was observed in the second real world experiment (\tabref{tab:results}.2).
Using the pushing, the offset distance was reduced by \SI{13.22}{mm} on average.
In the third real-world experiment (\tabref{tab:results}.3), the displacements, resulting from the orientation error, were smaller than the ones resulting from the translation error.
On average a displacement reduction of \SI{3.38}{mm} was achieved after pushing.

\subsubsection{Grasp Pose Approach:} 
\label{subsubsec:berscheid}

In the fourth experiment (\figref{fig:errorplots}.4 and \tabref{tab:results}.4), it is evaluated how our approach can deal with errors resulting from using the grasp pose estimator from Berscheid \etal~\cite{berscheidRobotLearningDoF2021a}.
As in the previous experiments, a significant reduction in the pose displacement was achieved.
In simulation (\tabref{tab:results}.4) a mean reduction of \SI{4.77}{mm} was measured.
In real world, the reduction was about three times larger (\SI{13.11}{mm}).
These values are similar to the ones measured in the second experiment (\tabref{tab:results}.2) and show that our error correction approach is applicable in such a scenario.

\subsection{Qualitative Evaluation of Exemplary Arrangement}

\begin{figure}[t]
	\centering
	\includegraphics[width=0.9\textwidth]{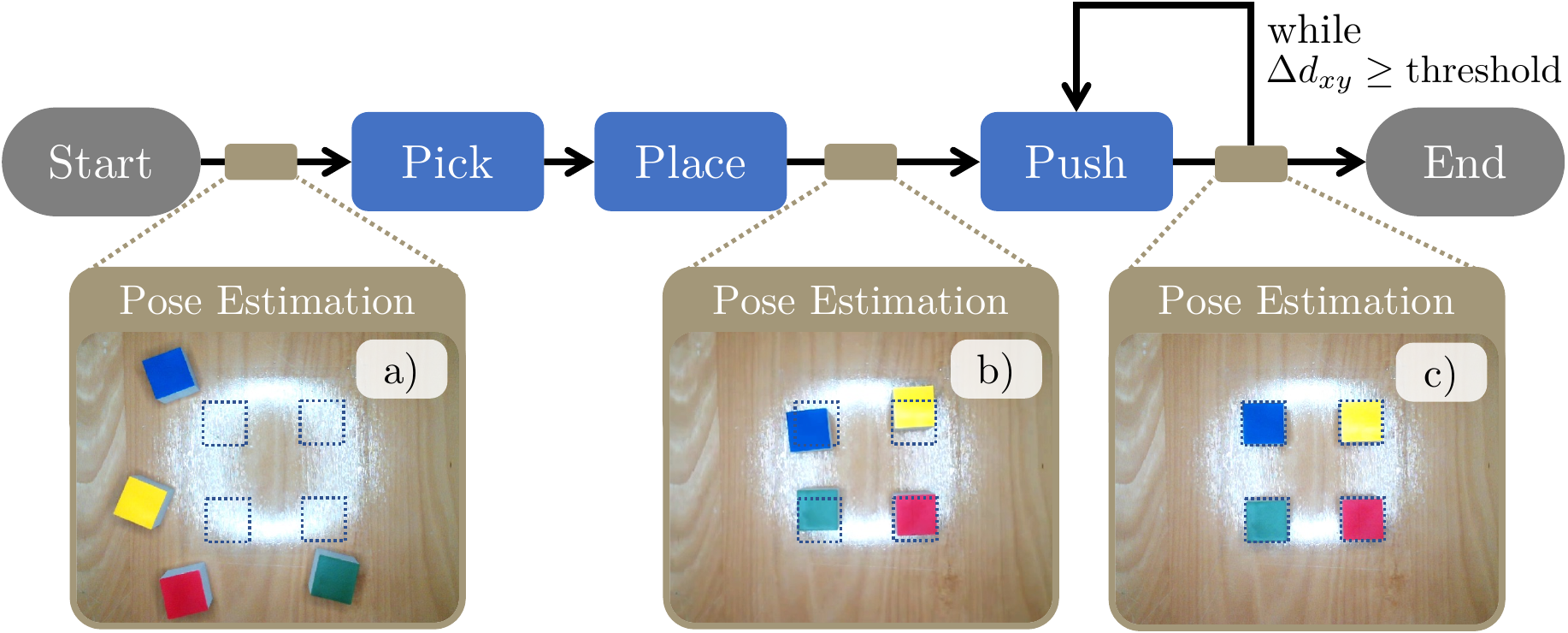}
	\caption{The images taken with the top camera during the arrangement experiment show the object poses before picking (a), after placing (b), and after pushing (c). Misplacements after the placing were completely eliminated by the pushing.}
	\label{fig:experiment}
\end{figure}

The last experiment was conducted to prove the applicability of our approach in an exemplary arrangement using four cubes with different colors.
For this experiment \algref{algo:flow} was altered so that the pushing is performed after placing all cubes to allow the reader the visual verification of the whole pattern.
\figref{fig:experiment}\,a shows the objects before picking the cubes.
They are scattered around the manipulation area.
During the pick a random translation and orientation error is applied to the gripper to simulate a picking error leading to the object's misplacement (\figref{fig:experiment}\,b).
After the final correction step (\figref{fig:experiment}\,c) all cubes are arranged in the desired pattern.
This demonstrates that our approach is able to accurately place cubes with a final mean error of less than \SI{1}{mm}. 

\section{Conclusion}
\label{sec:conclusion}
In this paper, we first presented a systematic categorization of manipulation errors that can cause object misplacements.
We then proposed a novel approach to reactively correct misplaced objects.
Furthermore, we demonstrated that our approach of iterative pushing leads to a significant improvement in the final placement pose in both simulation and the real world.
In addition to the quantitative experiments, we also showed the applicability of our approach for arranging objects using an exemplary pattern of colored cubes using a real robotic system.
As the experimental evaluation demonstrates, it is crucial to carry out reactive correction actions to obtain accurate object arrangements.
The arrangement accuracy is limited by physical boundaries of the pushing dynamics and the object pose estimation deriving, e.g., from the camera resolution.
In the future, we plan to extend the approach to cuboids and irregular objects using shape approximation techniques and we plan to apply it to the reconstruction of frescoes.

\section*{Acknowledgments}
\footnotesize{
We thank Patrick Reitz for helping with the experiments and Camila Maslatón for supporting with \figref{fig:motivation} and \ref{fig:manipulationErrors}.
This work has partly been supported by the RePAIR project of the European Union’s HORIZON 2020 research and innovation program under grant agreement n°964854.
} 


\bibliographystyle{splncs03}
\bibliography{bibliography}

\end{document}